\documentclass[journal]{IEEEtai}

\usepackage[colorlinks,urlcolor=blue,linkcolor=blue,citecolor=blue]{hyperref}

\usepackage{color,array}

\usepackage{graphicx}

\usepackage{amsmath,amsfonts}
\usepackage{textcomp}
\usepackage{stfloats}
\usepackage{url}
\usepackage{verbatim}
\usepackage{balance}

\usepackage{graphics} 
\usepackage{epsfig} 
\usepackage{mathptmx} 
\usepackage{times} 
\usepackage{amsmath} 
\usepackage{amssymb}  
\usepackage{siunitx}
\usepackage{hyperref}
\usepackage{multirow}
\usepackage{multicol}
\usepackage{nicefrac}
\usepackage{makecell}
\usepackage{svg}
\usepackage{cite}
\usepackage{picinpar}
\usepackage[latin1]{inputenc}
\usepackage{colortbl}
\usepackage{soul}
\usepackage{multirow}
\usepackage{pifont}
\usepackage{color}
\usepackage{alltt}
\usepackage{enumerate}
\usepackage{siunitx}
\usepackage{epstopdf}
\usepackage{pbox}
\usepackage{amsfonts}
\usepackage{textcomp}
\usepackage{diagbox}
\usepackage{graphics} 
\usepackage{times} 
\usepackage{amssymb}  
\usepackage{microtype}
\usepackage{subfigure}
\usepackage{booktabs} 
\usepackage{xfrac}
\usepackage{nicefrac}
\usepackage{multicol}
\usepackage{float}
\usepackage{array}
\usepackage{algorithm}
\usepackage{gensymb}
\usepackage{xr}
\usepackage{algpseudocode}
\usepackage{orcidlink}

\begin{document}

\title{Memory-Augmented Architecture for Long-Term Context Handling in Large Language Models} 

\author{
        Haseeb Ullah Khan Shinwari
 	and
    	Muhammad Usama \orcidlink{0000-0002-3834-2167} \IEEEauthorrefmark{1}
 
	\thanks{
            Haseeb Ullah Khan Shinwari is with Newton AI Lab (email: mr.haseebe@gmail.com)

            Muhammad Usama is with the School of Electrical Engineering, Korea Advanced Institute of Science and Technology (KAIST), Daejeon 34141, Republic of Korea (email: usama@kaist.ac.kr).
            
            Both authors contributed equally to this work.
            
 	}
  \thanks{\IEEEauthorrefmark{1}Corresponding author}
}


\markboth{Journal of IEEE Transactions on Artificial Intelligence, Vol. 00, No. 0, Month 2020}
{First A. Author \MakeLowercase{\textit{et al.}}: Bare Demo of IEEEtai.cls for IEEE Journals of IEEE Transactions on Artificial Intelligence}

\maketitle

\begin{abstract}
Large Language Models face significant challenges in maintaining coherent interactions over extended dialogues due to their limited contextual memory. This limitation often leads to fragmented exchanges and reduced relevance in responses, diminishing user experience. To address these issues, we propose a memory-augmented architecture that dynamically retrieves, updates, and prunes relevant information from past interactions, ensuring effective long-term context handling. Experimental results demonstrate that our solution significantly improves contextual coherence, reduces memory overhead, and enhances response quality, showcasing its potential for real-time applications in interactive systems.
\end{abstract}

\begin{IEEEImpStatement}
Maintaining long-term context is a persistent challenge for large language models (LLMs) in extended dialogues, often resulting in fragmented and incoherent responses. Our proposed memory-augmented architecture introduces a modular, relevance-based pruning mechanism that efficiently manages memory while preserving essential contextual information. This method enhances dialogue coherence, reduces memory overhead, and operates without significant architectural modifications, making it adaptable to various LLM frameworks.

The proposed solution significantly improves real-time applications such as virtual assistants, chatbots, and customer support systems by enabling sustained, contextually aware interactions. Furthermore, it reduces computational demands, ensuring scalability for deployment in resource-constrained environments. This advancement not only raises the standard for dialogue systems but also opens new avenues for AI applications requiring robust long-term memory management, bridging the gap between human-like conversational fluency and technological feasibility.
\end{IEEEImpStatement}



\section{Introduction}
Large Language Models (LLMs) have made significant strides in natural language processing (NLP) applications, powering chatbots, virtual assistants, and customer service agents. Despite their success, a critical challenge remains: maintaining long-term context across multiple interactions. Standard LLMs process each query independently, often overlooking the continuity of prior exchanges. This limitation can result in responses that lack coherence, contextual relevance, and even lead to hallucinations in extended dialogues or complex tasks \cite{hallucinations1, hallucinations2}.

The fixed input size of transformer-based architectures further exacerbates the issue by limiting the ability to capture long-term dependencies. As conversations grow, important contextual information is lost, leading to fragmented exchanges and a diminished user experience. While existing memory-augmented approaches attempt to store past interactions, they often face memory bloat and increased inference latency, making them impractical for real-time applications. Therefore, there is a pressing need for an efficient architecture that retrieves only the most relevant information while dynamically managing memory to maintain performance.

To address this challenge, various memory-augmented approaches have been proposed. These methods typically involve incorporating additional memory modules into the LLM architecture to store and retrieve relevant past information. For instance, LongMem \cite{wang2023augmentinglanguagemodelslongterm} utilizes a decoupled memory architecture with a retrieval network to select relevant memories. CAMELoT \cite{camelot} introduces an associative memory module to improve context handling within LLMs. However, these approaches often rely on LRU eviction for memory management, which may not be optimal for retaining highly relevant information.

Larimar \cite{larimar} is another recent approach that focuses on fast and efficient learning of contextual information. While Larimar shares the goal of improving context-awareness, it differs from your work in its focus on memory updates and adaptation rather than long-term retrieval.

\begin{figure*}[ht]
    \centering
    \includegraphics[width=\linewidth]{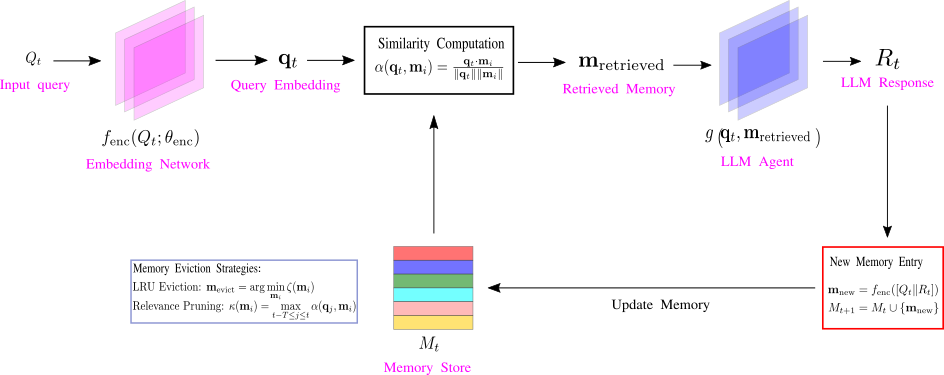}
    \caption{Context-Grounded Response Generation Framework for LLMs. This figure illustrates the proposed memory-augmented architecture designed for effective long-term context handling in LLMs. User queries $Q_t$ are mapped to embeddings $\textbf{q}_t$ via an embedding network, which are then matched with past interaction embeddings $\textbf{m}_i, \forall i$ in the memory store based on cosine similarity $\alpha(\textbf{q}_t, \textbf{m}_i)$. The most relevant memory $m_{\text{retrieved}}$ is selected to condition the LLM response $R_t$, grounding it in prior context. After generating the response, a new memory entry $\textbf{m}_{\text{new}}$ is added. To keep memory scalable, either Least Recently Used (LRU) eviction or relevance-based pruning is applied, ensuring the retention of only the most pertinent past interactions.}
    \label{fig:illustration}
\end{figure*}

In this paper, we propose a memory-augmented architecture, visualized in Figure \ref{fig:illustration}, that focuses on efficient long-term context handling. Unlike existing methods, we introduce a relevance-based pruning strategy to selectively retain the most pertinent information from past interactions. This approach complements LRU eviction by ensuring that crucial context is preserved, even if it's not the most recently accessed.

Furthermore, our method is designed to be modular and easily integrated with existing LLM architectures. By leveraging a retrieval network to select relevant memories, we avoid the need for significant modifications to the LLM itself. This flexibility allows for seamless integration with various LLM models and facilitates experimentation with different memory management strategies. 

Our contributions can be summarized as follows:
\begin{enumerate}
    \item Relevance-Based Pruning: We introduce a memory management strategy that prioritizes the retention of highly relevant information, improving the quality and coherence of LLM responses.

    \item Modular Architecture: Our approach is designed to be easily integrated with existing LLM architectures, requiring minimal modifications to the base model.

    \item Effective Long-Term Context Handling: Our method demonstrates improved performance in maintaining long-term context, leading to more coherent and relevant responses in dialogue systems.
\end{enumerate}

The proposed memory-augmented architecture differs from Retrieval-Augmented Generation (RAG) \cite{rag} by focusing on adaptive, real-time memory management tailored for ongoing dialogues, as opposed to RAG's static knowledge retrieval from fixed databases. Unlike RAG, which supplements isolated responses with external data, this approach dynamically updates, retrieves, and prunes memory entries based on contextual relevance, preserving interaction-specific information critical to long-term coherence. By employing relevance-based pruning instead of LRU-based eviction, the proposed method retains essential context efficiently, avoids memory bloat, and ensures scalability, making it well-suited for applications requiring sustained context adaptation over extended interactions.

\section{Context Grounded Response Generation in LLMs}
This section presents a memory-augmented framework designed for context-grounded response generation in Large Language Models (LLMs). The key challenge in maintaining coherent responses across long-term interactions lies in effectively managing the context accumulated from prior exchanges. Our approach addresses this challenge by introducing an adaptive memory mechanism that dynamically retrieves, updates, and manages relevant past interactions to inform future responses. 

Given a sequence of user inputs \( \{Q_1, Q_2, \dots, Q_t\} \), the goal is to generate a response \( R_t \) at time \( t \) that reflects both the current query \( Q_t \) and the pertinent context from prior interactions. To achieve this, each query is mapped to a high-dimensional embedding space using an encoder network \( f_{\text{enc}} \), parameterized by \( \theta_{\text{enc}} \):
\[
\mathbf{q}_t = f_{\text{enc}}(Q_t; \theta_{\text{enc}}).
\]
These query embeddings are then used to retrieve relevant memory entries from the memory store, ensuring that the responses generated by the LLM remain contextually grounded and coherent throughout the interaction.

The memory store at time \( t \), denoted by \( M_t = \{\mathbf{m}_1, \mathbf{m}_2, \dots, \mathbf{m}_N\} \), is a set of memory slots, each encoding information from previous interactions. To maintain relevance, a query-driven retrieval mechanism computes the similarity between the current query embedding \( \mathbf{q}_t \) and each memory entry \( \mathbf{m}_i \). The similarity score is defined as the cosine similarity:
\[
\alpha(\mathbf{q}_t, \mathbf{m}_i) = \frac{\mathbf{q}_t \cdot \mathbf{m}_i}{\|\mathbf{q}_t\| \|\mathbf{m}_i\|}.
\]
The memory entry with the highest similarity score is retrieved and used to condition the LLM's response generation process. The generated response \( R_t \) is expressed as:
\[
R_t = g(\mathbf{q}_t, \mathbf{m}_{\text{retrieved}}; \theta_{\text{dec}}),
\]
where \( g \) is the decoder network of the LLM parameterized by \( \theta_{\text{dec}} \). 

After generating the response, the memory store is updated with the new interaction. The new memory entry is constructed by encoding the concatenated query and response:
\[
\mathbf{m}_{\text{new}} = f_{\text{enc}}([Q_t \Vert R_t], \theta_{\text{enc}}).
\]
The updated memory store becomes:
\[
M_{t+1} = M_t \cup \{\mathbf{m}_{\text{new}}\}.
\]

Since the memory size grows with each interaction, efficient memory management is crucial to ensure the system remains scalable. We introduce two memory management strategies: Least Recently Used (LRU) eviction and relevance-based pruning. In the LRU strategy, the memory entry with the smallest access time is evicted to make room for new entries:
\[
\mathbf{m}_{\text{evict}} = \arg\min_{\mathbf{m}_i \in M_t} \zeta(\mathbf{m}_i),
\]
where \( \zeta(\mathbf{m}_i) \) is the timestamp of the last access of memory \( \mathbf{m}_i \). In relevance-based pruning, a relevance score \( \kappa(\mathbf{m}_i) \) is assigned to each memory slot based on the maximum similarity with recent queries:
\[
\kappa(\mathbf{m}_i) = \max_{t - T \leq j \leq t} \alpha(\mathbf{q}_j, \mathbf{m}_i),
\]
where \( T \) is the window size of recent queries. The least relevant memory entry is removed:
\[
\mathbf{m}_{\text{prune}} = \arg\min_{\mathbf{m}_i \in M_t} \kappa(\mathbf{m}_i).
\]
These strategies prevent the memory store from growing indefinitely, ensuring that only the most relevant interactions are retained.

The complete algorithm for long-term context handling with memory management is presented in Algorithm~\ref{alg:ltc_handling}.

\begin{algorithm}[t]
\caption{Long-Term Context Handling with Memory Management}
\label{alg:ltc_handling}
\begin{algorithmic}[1]

\State \textbf{Initialize:} Memory $M_0 = \emptyset$; Time $t = 1$; Max size $N$
\While {New query $Q_t$ arrives}
    \State \textbf{Embed Query:} $\mathbf{q}_t = f_{\text{enc}}(Q_t; \theta_{\text{enc}})$
    
    \State \textbf{Retrieve Memory:}\\
    \[
    \mathbf{m}_{\text{retrieved}} = \arg\max_{\mathbf{m}_i \in M_t} \alpha(\mathbf{q}_t, \mathbf{m}_i),
    \]
    \[ \text{where} \quad
    \alpha(\mathbf{q}_t, \mathbf{m}_i) = \frac{\mathbf{q}_t \cdot \mathbf{m}_i}{\|\mathbf{q}_t\| \|\mathbf{m}_i\|}
    \]

    \State \textbf{Generate Response:} $R_t = g(\mathbf{q}_t, \mathbf{m}_{\text{retrieved}}; \theta_{\text{dec}})$

    \State \textbf{Update Memory:} $\mathbf{m}_{\text{new}} = f_{\text{enc}}([Q_t \Vert R_t], \theta_{\text{enc}})$
    \State $M_{t+1} \gets M_t \cup \{\mathbf{m}_{\text{new}}\}$

    \If {Memory size exceeds $N$}
        \State \textbf{Manage Memory:}
        \State \textbf{Option 1: LRU Eviction} 
        \[
        \mathbf{m}_{\text{evict}} = \arg\min_{\mathbf{m}_i \in M_t} \zeta(\mathbf{m}_i)
        \]
        \State $M_{t+1} \gets M_t \setminus \{\mathbf{m}_{\text{evict}}\} \cup \{\mathbf{m}_{\text{new}}\}$
        
        \State \textbf{Option 2: Relevance-Based Pruning}
        \State Compute $\kappa(\mathbf{m}_i) = \max_{t - T \leq j \leq t} \alpha(\mathbf{q}_j, \mathbf{m}_i)$
        \State $\mathbf{m}_{\text{prune}} = \arg\min_{\mathbf{m}_i \in M_t} \kappa(\mathbf{m}_i)$
        \State $M_{t+1} \gets M_t \setminus \{\mathbf{m}_{\text{prune}}\} \cup \{\mathbf{m}_{\text{new}}\}$
    \EndIf

    \State $t \gets t + 1$
\EndWhile

\end{algorithmic}
\end{algorithm}

\section{Experiments and Results}
This section provides a comprehensive analysis of our memory-augmented model's performance on various tasks. 

\subsection{Experimental Setup}
We evaluate our proposed memory-augmented framework on three tasks known for challenging a model's memory and contextual capabilities:
\begin{itemize}
    \item 20 Questions Game: This turn-based guessing game requires models to infer an unknown entity within 20 questions based on hints provided during each turn. Given the cumulative nature of the clues, efficient memory handling and relevance-driven filtering are critical.
    \item Persona-Chat: This dataset contains dialogues where each agent follows a predefined persona. It emphasizes the importance of memory consistency across turns, as the model must maintain persona coherence while responding contextually and appropriately.
    \item DailyDialog: DailyDialog comprises over 13,000 dialogues across various topics, providing a natural conversational setting that tests a model's ability to retain and leverage long-term context for more realistic conversational flows.
\end{itemize}

To ensure fair comparisons, both baseline and memory-augmented variants of LLAMA 3 8B and Gemma 2 9B are tested under identical hardware setups: an Intel Xeon 3.4 GHz processor, 64 GB RAM, and an NVIDIA RTX 3090 GPU, with PyTorch \cite{pytorch} used for model training and testing. We use GTE-large \cite{gte-large} as the embedding network in our experiments. 

\subsection{Tasks and Datasets}
We select a combination of synthetic and real-world datasets that pose unique memory-related challenges:

\subsubsection{20 Questions Game Dataset}
We created a custom dataset comprising 100 entities across ten categories, ranging from animals to objects. The model's objective is to deduce the correct entity within 20 turns, based on cumulative user hints.

\subsubsection{Persona-Chat Dataset}
This dataset contains dialogues between two agents, each equipped with a unique persona, requiring the model to recall persona-relevant details and respond with consistency over multiple turns.

\subsubsection{DailyDialog Dataset}
Comprising a wide variety of everyday topics, DailyDialog allows us to test the model's general conversation abilities and long-term memory retention in a natural conversational setting.

\subsection{20 Questions Game Environment}

The 20 Questions game environment simulates an interactive guessing game where a player, designated as the \textit{guesser}, attempts to identify a secret keyword through a series of yes-or-no questions directed at an \textit{answerer}. The environment is structured to foster a dialogue between two agents, which can either be LLM-based or random guessers and answerers, allowing for a controlled assessment of the LLM's capabilities in handling long-term context and maintaining coherence in dialogue.

At the start of each game, a keyword is randomly selected from a predefined list of categories and associated keywords. This keyword represents a real or fictional entity that the guesser aims to identify. The environment employs a set of structured prompts to facilitate communication between the guesser and answerer agents. When the guesser asks a question, the answerer provides a response based on the current state of the game, which may include ``yes,'' ``no,'' or ``maybe,'' depending on the nature of the question posed.

The game progresses through alternating turns between the guesser and answerer, with the state of the game captured in observations that include the questions asked, the answers provided, and any guesses made. Each agent maintains a record of their interactions, which is crucial for both short-term decision-making and long-term contextual awareness. This design emphasizes the necessity for the agents to recall past exchanges to make informed guesses and responses, directly testing the memory capabilities of the LLM.

To manage the flow of the game, a turn-based mechanism is implemented where agents alternate roles. If the guesser successfully identifies the keyword within the allowed number of questions, the game concludes positively for that agent. Conversely, if the guesser fails to guess the keyword within the stipulated turns, the game ends in failure. The environment also incorporates mechanisms to handle erroneous responses and timeouts, ensuring that the game can adapt to various scenarios and maintain its integrity.


The algorithm showcasing the working of the 20-questions game environment is given in Algorithm \ref{alg:20_questions} and the code implementation is given at \cite{kaggle_env}.

\begin{table*}[ht]
\centering
\caption{Performance on the 20 Questions Game}
\label{tab:20q_results}
\begin{tabular}{l|c|c|c|c|c}
\hline
\textbf{Model} & \textbf{Method} & \textbf{Accuracy (\%)} & \textbf{Latency (ms)} & \textbf{Memory Overhead (MB)} & \textbf{PTR (\%)} \\
\hline
\multirow{2}{*}{LLAMA 3 8B} & Baseline & 62.3 & - & - & 0.0 \\
& Proposed & 80.4 & 1287.80 & 1022.7 & 35.2 \\
\hline
\multirow{2}{*}{Gemma 2 9B} & Baseline & 64.8 & - & - & 0.0 \\
& Proposed & 82.1 & 1137.3 & 1173.5 & 36.8 \\
\hline
\end{tabular}
\end{table*}

\begin{algorithm}[ht]
    \caption{20 Questions Game Environment}
    \label{alg:20_questions}
    \begin{algorithmic}[1]
        \State Initialize the game environment
        \State Select a random keyword and category
        \State Set turn = 1
        \State Set maximum turns = 20
        \While{turn $\leq$ maximum turns}
            \If{guesser's turn}
                \State prompt the guesser for a question
                \State record the question
                \State send question to answerer
                \State receive answer (yes/no/maybe)
                \If{guesser makes a guess}
                    \If{guess is correct}
                        \State End game with success
                    \Else
                        \State record the guess
                    \EndIf
                \EndIf
            \Else
                \State prompt the answerer for a response
                \State provide the answer to the guesser
            \EndIf
            \State Increment turn
        \EndWhile
        \If{maximum turns reached}
            \State End game with failure
        \EndIf
    \end{algorithmic}
\end{algorithm}


\subsection{Evaluation Metrics}
We employ the following metrics to evaluate the performance and memory efficiency of the proposed framework:

\subsubsection{Accuracy (20 Questions Game)}
 Measures the proportion of correct guesses made by the model, reflecting its ability to maintain and leverage context.
    \subsubsection{Contextual Coherence Score (CCS)} CCS measures the cosine similarity between each model response and the aggregated context vector:
    \[
    \text{CCS} = \frac{1}{N} \sum_{i=1}^{N} \text{sim}(r_i, C_i)
    \]
    where \(N\) is the number of turns, \(r_i\) the model's response, \(C_i\) the context vector, and \(\text{sim}(r_i, C_i)\) the cosine similarity between the response and the context.
    \subsubsection{Memory Overhead (MB)} Reflects the memory consumption throughout the interaction, providing insight into the model's scalability.
    \subsubsection{Latency (ms)} Records average response time per query, relevant for real-time applications.
    \subsubsection{Positive Transferability Ratio (PTR)} Quantifies the percentage of interactions where the memory-augmented model outperforms the baseline in response relevance:
    \[
    \text{PTR} = \frac{1}{M} \sum_{j=1}^{M} \mathbb{I}\left( \text{sim}(r_j^{\text{mem}}, C_j) > \text{sim}(r_j^{\text{base}}, C_j) \right)
    \]
    Here, \(M\) is the total interactions, \(r_j^{\text{mem}}\) the memory-augmented model response, \(r_j^{\text{base}}\) the baseline response, and \(C_j\) the interaction context.

\subsection{Results and Analysis}

\subsubsection{20 Questions Game}
Table~\ref{tab:20q_results} presents the accuracy, latency, memory overhead, and PTR for both models on the 20 Questions game. The memory-augmented versions of LLAMA 3 8B and Gemma 2 9B show substantial accuracy gains over the baseline, with LLAMA 3 8B improving from 62.3\% to 80.4\% and Gemma 2 9B from 64.8\% to 82.1\%. This accuracy increase underscores the proposed memory mechanism's effectiveness in maintaining and retrieving relevant contextual information for accurate guesswork.

\subsubsection{Persona-Chat and DailyDialog}
Tables~\ref{tab:dialog_results} presents the CCS and PTR results for the Persona-Chat and DailyDialog tasks. The memory-augmented versions of both LLAMA 3 8B and Gemma 2 9B demonstrate higher CCS and PTR scores, indicating improved contextual coherence and relevance. Notably, Gemma 2 9B achieved the highest CCS (0.83) on Persona-Chat and PTR scores on both datasets, suggesting that larger models may benefit more significantly from memory augmentation in context-rich scenarios.

\begin{table}[h]
\centering
\caption{Performance on Dialogue Datasets}
\label{tab:dialog_results}
\begin{tabular}{l|l|c|c|c}
\hline
\textbf{Dataset} & \textbf{Model} & \textbf{Method} & \textbf{CCS} & \textbf{PTR (\%)} \\
\hline
\multirow{4}{*}{Persona-Chat} & \multirow{2}{*}{LLAMA 3 8B} & Proposed & 0.74 & 28.1 \\
& & Baseline & 0.65 & 25.0 \\
\cline{2-5}
& \multirow{2}{*}{Gemma 2 9B} & Proposed & 0.83 & 30.5 \\
& & Baseline & 0.72 & 27.5 \\
\hline
\multirow{4}{*}{DailyDialog} & \multirow{2}{*}{LLAMA 3 8B} & Proposed & 0.69 & 31.4 \\
& & Baseline & 0.60 & 29.0 \\
\cline{2-5}
& \multirow{2}{*}{Gemma 2 9B} & Proposed & 0.81 & 33.2 \\
& & Baseline & 0.75 & 30.0 \\
\hline
\end{tabular}
\end{table}

\subsection{Memory Management Impact}
We evaluated various memory management strategies, specifically relevance-based pruning and LRU eviction, on both the LLAMA 3 8B and Gemma 2 9B models. The assessment focused on key performance metrics such as accuracy, memory usage, and latency, as summarized in Table~\ref{tab:memory_management}, using the 20 Questions game task.

For the LLAMA 3 8B model, relevance-based pruning achieved the highest accuracy at 80.4\%, representing a notable increase from 78.1\% observed under the no pruning condition. This accuracy improvement is accompanied by a reduction in memory overhead (1022.7 MB compared to 1250.0 MB) and a slight decrease in latency (1287.8 ms versus 1400.3 ms). In contrast, the LRU eviction method resulted in the lowest accuracy at 76.5\%, highlighting its inadequacy in preserving relevant contextual information.

Similarly, the Gemma 2 9B model exhibited comparable trends. The no-pruning strategy provided the highest accuracy at 82.6\%, while relevance-based pruning closely followed with an accuracy of 82.1\%. This approach not only maintained efficient memory usage (1173.5 MB) but also resulted in significantly lower latency (1137.3 ms). Again, the LRU eviction strategy fell short in terms of accuracy (78.2\%), reinforcing the conclusion that relevance-based pruning is superior for effective contextual retention.

\begin{table*}[ht]
\centering
\caption{Impact of Memory Management Strategies on LLAMA 3 8B and Gemma 2 9B Models}
\label{tab:memory_management}
\begin{tabular}{l|l|c|c|c}
\hline
\textbf{Model} & \textbf{Strategy} & \textbf{Accuracy (\%)} & \textbf{Memory Overhead (MB)} & \textbf{Latency (ms)} \\
\hline
\multirow{3}{*}{LLAMA 3 8B} & No Pruning & 78.1 & 1250.0 & 1400.3 \\
& Relevance-Based Pruning & 80.4 & 1022.7 & 1287.8 \\
& LRU Eviction & 76.5 & 1030.4 & 1295.6 \\
\hline
\multirow{3}{*}{Gemma 2 9B} & No Pruning & 82.6 & 1320.5 & 1375.2 \\
& Relevance-Based Pruning & 82.1 & 1173.5 & 1137.3 \\
& LRU Eviction & 78.2 & 1103.4 & 1261.4 \\
\hline
\end{tabular}
\end{table*}

\subsection{Impact of Embedding Models}
The embedding network plays a crucial role in our architecture as it maps queries and responses to a high-dimensional space where similarity computations guide memory retrieval. We evaluated three widely used embedding models: \textit{GTE-large} \cite{gte-large}, Sentence Transformers \textit{MiniLM-L6-v2} \cite{mini} and \textit{Universal Sentence Encoder (USE)} \cite{Cer2018UniversalSE}. We tested these embedding models with both LLAMA 3 8B and Gemma 2 9B on the 20 Questions game task, maintaining all other architectural components constant.

\begin{table*}[t]
\centering
\caption{Impact of Different Embedding Models on 20 Questions Game Performance}
\label{tab:embedding_comparison}
\begin{tabular}{l|l|c|c|c}
\hline
\textbf{Model} & \textbf{Embedding Model} & \textbf{Accuracy (\%)} & \textbf{Memory Overhead (MB)} & \textbf{Latency (ms)} \\
\hline
\multirow{3}{*}{LLAMA 3 8B} 
& GTE-large & 80.4 & 1022.7 & 1287.8 \\
& MiniLM-L6-v2 & 75.8  & 985.3 & 1156.4 \\
& USE & 73.2 & 1012.5 & 1245.6 \\
\hline
\multirow{3}{*}{Gemma 2 9B} 
& GTE-large & 82.1 & 1173.5 & 1137.3 \\
& MiniLM-L6-v2 & 77.3 & 1125.8 & 1089.5 \\
& USE & 74.9 & 1158.2 & 1112.7 \\
\hline
\end{tabular}
\end{table*}

Table~\ref{tab:embedding_comparison} presents the results of this analysis. The results demonstrate that the choice of embedding model significantly impacts the system's performance. GTE-large consistently outperforms other embedding models across both LLAMA 3 8B and Gemma 2 9B, achieving the highest accuracy scores (80.4\% and 82.1\% respectively). This superior performance can be attributed to GTE-large's enhanced ability to capture semantic relationships and nuanced contextual information critical for the 20 Questions game.

MiniLM-L6-v2, while more computationally efficient as evidenced by lower latency times, shows a moderate decrease in accuracy (75.8\% for LLAMA 3 8B and 77.3\% for Gemma 2 9B). The Universal Sentence Encoder, despite its general-purpose design, achieves the lowest accuracy scores among the three models (73.2\% and 74.9\% respectively).

\section{Discussion}

Our evaluation of the memory-augmented context-handling framework provides valuable insights into model performance, contextual retention, and memory efficiency across diverse tasks. In the 20 Questions game, both the LLAMA 3 8B and Gemma 2 9B models showed substantial improvements in accuracy with memory augmentation, underscoring the framework's ability to effectively maintain and leverage context. Specifically, LLAMA 3 8B improved from 62.3\% to 80.4\%, and Gemma 2 9B advanced from 64.8\% to 82.1\%, illustrating the significant role of memory in improving the accuracy of sequential task performance.

Further analysis on the Persona-Chat and DailyDialog datasets revealed that memory augmentation also enhances contextual coherence and relevance, as measured by the Contextual Coherence Score (CCS) and Positive Transferability Ratio (PTR). Gemma 2 9B achieved the highest CCS of 0.83 on Persona-Chat, with consistently higher PTR scores across both datasets. These results suggest that larger models, like Gemma 2 9B, may benefit even more from memory mechanisms, especially in tasks with rich contextual elements, where maintaining coherent dialogue over multiple turns is critical.

In examining memory management strategies, relevance-based pruning emerged as the more effective approach compared to LRU eviction. For LLAMA 3 8B, relevance-based pruning achieved an accuracy of 80.4\%, while reducing memory overhead to 1022.7 MB and latency to 1287.8 ms, compared to no pruning. Gemma 2 9B followed a similar trend, with relevance-based pruning achieving 82.1\% accuracy, while efficiently reducing memory usage to 1173.5 MB and lowering latency to 1137.3 ms. By contrast, LRU eviction led to the lowest accuracies of 76.5\% for LLAMA 3 8B and 78.2\% for Gemma 2 9B, highlighting its limitations in retaining essential contextual information. These findings reinforce the advantage of relevance-based pruning, particularly for tasks requiring precise and coherent contextual retention.

Our ablation study demonstrated that the GTE-large embedding consistently provided the highest accuracy across both model variants, achieving 80.4\% for LLAMA 3 8B and 82.1\% for Gemma 2 9B.  This superior performance can be attributed to GTE-large's enhanced capacity for capturing semantic relationships and subtle contextual nuances essential for complex dialogue tasks.  In comparison, MiniLM-L6-v2 showed computational efficiency benefits with lower latency (1156.4 ms for LLAMA 3 8B and 1089.5 ms for Gemma 2 9B) but experienced a moderate decline in accuracy (75.8\% and 77.3\%, respectively). The Universal Sentence Encoder, while useful for general-purpose applications, yielded the lowest accuracy scores (73.2\% for LLAMA 3 8B and 74.9\% for Gemma 2 9B), indicating that specialized embedding models like GTE-large are better suited for tasks requiring deep contextual understanding.

\section{Conclusion}
In this paper, we propose a memory-augmented architecture to address the limitations of large language models in handling long-term context. Our approach incorporates a retrieval network to selectively access relevant past interactions and a relevance-based pruning strategy to optimize memory usage. Experimental results on various tasks, including the 20 Questions Game, Persona-Chat, and DailyDialog, demonstrate significant improvements in accuracy, contextual coherence, and response relevance. Our findings suggest that memory augmentation is a promising technique for enhancing the capabilities of LLMs in real-world dialogue systems.

\bibliographystyle{IEEEtran}
\bibliography{references}

\end{document}